\documentclass[conference]{IEEEtran}
\IEEEoverridecommandlockouts
\usepackage{cite}
\usepackage{amsmath,amssymb,amsfonts}
\usepackage{algorithmic}
\usepackage{graphicx}
\usepackage{textcomp}
\usepackage{xcolor}
\usepackage{array}
\def\BibTeX{{\rm B\kern-.05em{\sc i\kern-.025em b}\kern-.08em
    T\kern-.1667em\lower.7ex\hbox{E}\kern-.125emX}}
\begin{document}

\title{Improving LSTM Neural Networks for Better Short-Term Wind Power Predictions}
\author{\IEEEauthorblockN{Maximilian Du}
\IEEEauthorblockA{\textit{Fayetteville-Manlius High School} \\
Manlius, NY, United States\\
maxressb@gmail.com}}
\maketitle
\begin{abstract}
This paper improves wind power prediction via weather forecast-contextualized Long Short-Term Memory Neural Network (LSTM) models. Initially, only wind power data was fed to a generic LSTM, but this model performed poorly, with erratic and naive behavior observed on even low-variance data sections. To address this issue, weather forecast data was added to better contextualize the power data, and LSTM modifications were made to address specific model shortcomings. These models were tested through both a Normalized Mean Absolute Error and the Naive Ratio (NR), which is a score introduced by this paper to quantify the unwanted presence of naive character in trained models. Results showed an increased accuracy with the addition of weather forecast data on the modified models, as well as a decrease in naive character. Key contributions include making improved LSTM variants, usage of weather forecast data, and the introduction of a new model performance index.
\end{abstract}
\section{Introduction}
Wind energy is both a clean and renewable source of energy, and its usage across the world is increasing. However, because the wind is uncontrollable, wind turbine farms can cause grid instability, and the cost of cycling conventional power plants to compensate for this variability is around 157 million dollars per year in areas with only 35\% integration \cite{Lew}. In the future, this instability and subsequent costs will only worsen as wind power penetration increase and conventional power capacity decrease \cite{AWEA}. 

Predicting wind power output reduces these adverse effects of wind power penetration by playing important roles in load balancing \cite{Khorramdel} and reserve optimization\cite{Wang}. Furthermore, as conventional capacity decreases, more accurate predictions over ultra-short intervals of time (5-10 minutes) are needed to maintain power grid stability \cite{Riahy}.

There are many techniques available for time series problems, e.g. wind power prediction, such as mathematical and statistical modeling. However, machine learning approaches can be superior in terms of robustness and accuracy because they are very adaptable and do not rely on simulating a turbine environment \cite{Xiaoyun}. One such machine learning model is the Long Short-Term Memory Neural Network (LSTM). The LSTM is a variant of the Recurrent Neural Network (RNN) that uses a continuous cell state to carry temporal information. Such a cell state is modified by the Forget, Input/Input Transform, and Output gates to forget, add, and recall information, respectively. The LSTM was shown to be powerful at time-series prediction, and once trained, they are relatively lightweight  \cite{Hochreiter}. 

Existing research has already approached the problem of wind power prediction using LSTMs, e.g. Xiaoyun et al \cite{Xiaoyun}. However, while such research considered the importance of \emph{current weather data} from the Numerical Weather Prediction (NWP) model as contextualization, the usage of weather \emph{forecast data} also provided by the NWP remains as an additional and less-explored option.

This paper improves the accuracy of the LSTM model by using NWP forecast data. Furthermore, this paper proposes a new way of numerically quantifying the presence of \textit{Naive Character}, an undesirable inclination to directly use current states as predictions for future states. To decrease such character and increase prediction accuracy, multiple new LSTM modifications were designed, tested, and compared using a Normalized Mean Absolute Error (NMAE) as well as a newly proposed Naive Ratio.
\section{Motivational Example: Problems Encountered with the Generic LSTM}
As a starting point, unmodified (generic) LSTMs were trained and evaluated on power data using the Tensorflow Python library. Although LSTMs have been established as reliable models, there are several problems observed with the trained model. 
\subsection{Naivety and Step Plots}
Naive character, or the \emph{Persistence Algorithm}, is the undesirable usage of the current state as the prediction for the next state. This is mathematically represented as $x'_{t+1}=x_t$. On a step plot, which holds each point on the plot constant until the next point, naive character is observed as a one-step shift between truth and prediction. Other works such as Chang et al \cite{Chang} and Xiaoyun et al \cite{Xiaoyun} have generated prediction vs. truth line plots, but these line plots do not visibly show naive character, especially when viewed over a long time range relative to their timestep. 
\begin{figure}[htb] 
        \centering \includegraphics[width=0.7\columnwidth]{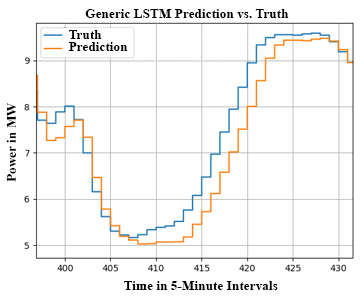}
                \caption{Step plot of Generic LSTM output shows Naive behavior, which indicates that this model has not learned any real trends.}
                        \label{fig:naive}
                        \end{figure}
In Figure \ref{fig:naive}, a step plot, this offset between the truth and prediction plots is apparent. The presence of this naive behavior on the generic LSTM shows that the trained model has found the persistence algorithm as a local minimum instead of truly modeling wind power trends. 
\subsection{Predication Oscillation and Negative Outputs}
Another problem observed through step plots and cell state distributions was prediction oscillation caused by cell state divergence. 
\begin{figure}[htb] 
        \centering \includegraphics[width=0.70\columnwidth]{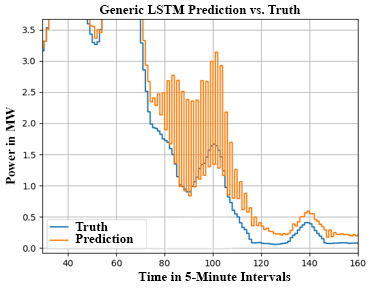}
                \caption{A different step plot section of Generic LSTM output shows prediction oscillation, which indicates that this model is unstable.}
                        \label{fig:oscilate}
                        \end{figure}
As can be seen in Figure \ref{fig:oscilate}, which shows a different section of the test set, the prediction values oscillate from point to point. This problem can be justified by the magnitudes of the cell states, whose outliers diverged up to $\pm 10^5$ at the end of training. With this range, optimization favored very small output weights to produce predictions of similar magnitude as the truth. These diminishing weights attenuate more common mid-range cell values, leading to under-sensitivity to specific states and over-sensitivity to others, resulting in unstable and inconsistent outputs.

Negative outputs, which are undesirable for power predictions, were also observed in this model due to the lack of output regulation. 
\section{Data Acquisition and Processing}
Data from two open-source databases was used for training and testing the models. 
\subsection{Wind Power Data}
This database from the National Renewable Energy Laboratories (NREL) \cite{windprospector} was chosen as the source of power data. For this paper, data from a 16 MW capacity turbine farm off the coast of \textit{Long Island, New York, United States} was used due to its relatively greater power fluctuations. 
\subsection{Weather Forecast Data}
It was hypothesized that weather forecast data would enable the model to have some future context while harnessing the power of existing weather models. This forecast data was obtained from the Rapid Refresh (RAP) database from NOAA \cite{RAP}. While there existed a large amount of weather information (317 parameters over 18 hours of forecast), to prevent overloading the LSTM with unnecessary amounts of data, specific weather parameters across an empirically chosen 2-hour forecast interval were used. These included the following:
\begin{itemize}
\item Surface air pressure
\item Ground temperature
\item 2-meter air temperature
\item Relative humidity
\item Wind gust peak velocities
\end{itemize}
All of the above parameters were chosen based on their ability to represent current and future weather conditions relevant to wind power generation.
\subsection{Data Concatenation and Implementation}
The NREL wind power set was found in five-minute increments, while the RAP dataset was found in one-hour increments. To combine them, the RAP dataset values were kept constant in blocks of 12 wind power points. 

The addition of weather forecast data to the power data meant increasing the input vector size.  As a starting point, a direct feed-in was used, where all input values were made into input nodes. This was done to determine the effectiveness of the forecast parameters in improving model performance. 

Compression algorithms like Principal Component Analysis (PCA), used in Xiaoyun et al \cite{Xiaoyun}, can reduce noise and complexity, so a model using PCA was made to gauge performance impacts.  It was observed that the first principle component contained more than 98\% of the variance, so it was the only component used. 

\section{Proposed Structure Modifications}
With the addition of weather data and the usage of PCA, there was a significant increase in the prediction accuracy of the generic LSTM (Table \ref{Generic}). However, the forecast data did not address prediction oscillation, and naive character was still observed, which indicated sub-optimal trend learning. Thus, some structural modifications were needed to address specific problems of the Generic LSTM and improve performance. 
\subsection{Main Modification: Cell Augmentation and Hyperbolic Tangent Addition}
\begin{figure}[htb] 
        \centering \includegraphics[width=0.9\columnwidth]{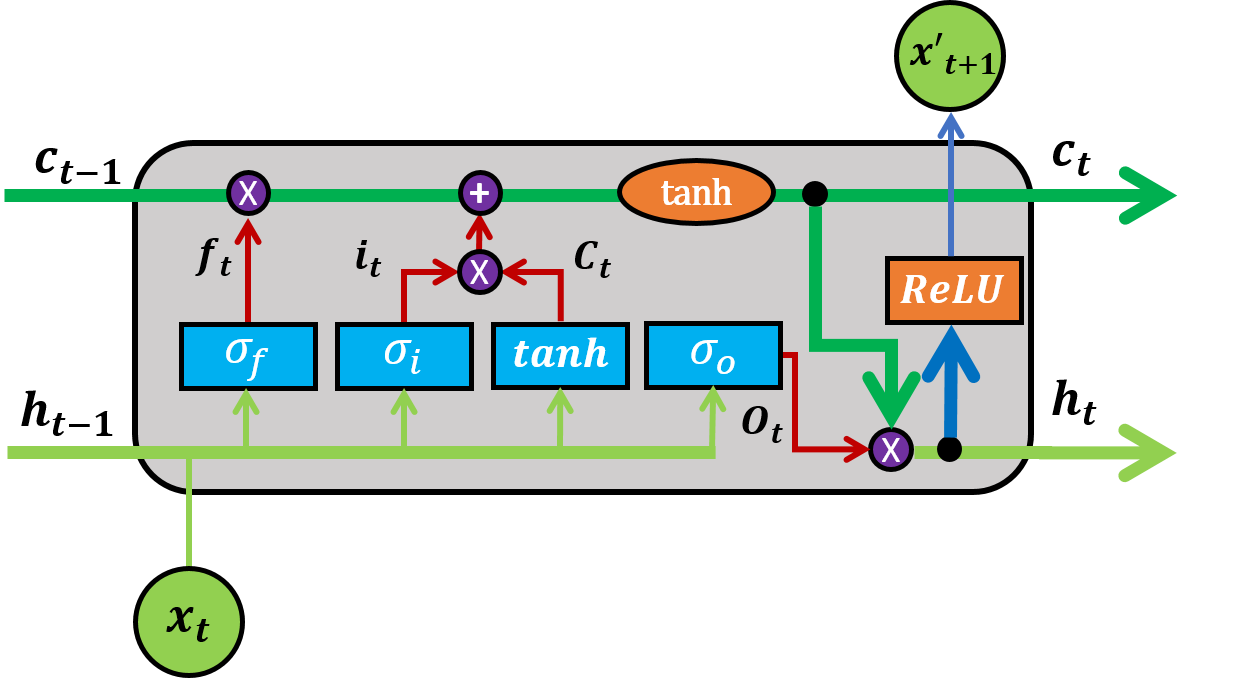}
                \caption{Model of LSTM with added hyperbolic tangent and cell augmentation, which greatly increased accuracy.}
                        \label{fig:tanh}
                        \end{figure}
To address prediction oscillation, a hyperbolic tangent activation function (shown as an orange oval in Figure \ref{fig:tanh}) was added to the cell propagation, which keeps values within the interval $(-1,1)$. This removed the cell divergence phenomenon found in the generic model, and oscillations were no longer observed. 

To further decrease naive character and to account for the increased amount of information, the cell state dimension of the LSTM was made independent of input size, allowing for the expansion of state sizes to keep more temporal information. To create this cell dimension independence, a fully-connected single-layer neural network with a rectified linear unit (ReLU) activation function was added after output gating (shown as an orange square in Figure \ref{fig:tanh}). This activation function eliminated negative outputs. 

Initial testing showed that these modifications significantly increased prediction accuracy (Table \ref{modelresults}), so the structure was kept as a backbone model, denoted \textbf{mLSTM}, or the \textit{modified LSTM}. 

Because the addition of the hyperbolic tangent re-introduced the vanishing gradient problem that prevents effective training on continuous data, \emph{hybrid backpropagation} was used. This method uses discrete blocks of sequential data for model training while initializing cell and hidden states with those from the previous iteration, allowing for some long-term trend learning without the use of an extended backpropagation. 
\subsection{Combined Input and Forget Gates (CIFG) with Peephole Connections}
The CIFG modification, explored in Greff et al \cite{Greff}, uses a single gate to modify both the Input Transform and Forget values, setting $f_t = 1-i_t$. Thus, the more attenuated a cell value becomes after forget gating, the more it is modified by the input gate. This increases memory efficiency by preventing excessive null (zero) values in the cell state.

The peephole, explored in Gers et al \cite{Gers}, gives the gates information about the current cell state and allows the LSTM to learn precise timings by enabling two-way data flow, which is beneficial for long-term trends and can reduce naive behavior. 

These modifications were proposed before, but there exists limited literature on their usage together and none on the new \emph{mLSTM}, yet it has been reported that both modifications have a positive impact on LSTM models. 
\subsection{Compression Layer}
Data compression in the form of PCA was previously used on the weather data to decrease input complexity. In order to gauge the effects of a trainable compression algorithm on the model, a compression layer was proposed for the \emph{mLSTM}. This modification takes the raw input vector and compresses it through a fully-connected layer. The construction is similar to that of an autoencoder, but it is trained directly with the model. Thus, the compression layer can learn to isolate application-relevant data components, which may or may not be the components that are the most important for data preservation. To keep results comparable with the PCA algorithm, which only used one principle value, the compression layer was also trained to compress down to a single scalar value. 
\section{Cross-Model Comparison}
\subsection{Hyperparameter Optimization}
The \emph{mLSTM} and its variants had three hyperparameters that were adjusted:
\begin{itemize}
    \item Learning rate
    \item Cell dimension
    \item Instance iteration length
\end{itemize}
The learning rate tunes the speed of gradient descent, the cell dimension sets the size of the cell and hidden states, and the instance iteration length determines the number of past data points fed to the model per iteration. A higher learning rate results in faster convergence, but it also increases the chance of sub-optimal endpoints. Likewise, a larger cell dimension increases the ability to store more information, but past a certain threshold, the extra space will only serve to increase complexity. Lastly, more data points given to the model leads to greater long-term trend recognition but it also increases the effects of the vanishing gradient problem, even with hybrid backpropagation. 

There are no guarantees that each of the LSTM modifications shares the same set of optimal hyperparameters, so a genetic hyperparameter optimization was run on each modification. The genetic algorithm created 10 child models and then partially trained each child to 3000 epochs. Each model was then evaluated on the same test set, and the hyperparameters of the best two children were kept. Ten new children were made from the two vectors, using crossover and mutation to increase variance and decrease the chance of getting stuck in a local minimum. This was done until stabilization of all values was achieved, which was empirically set at 20 generations for all models.
\subsection{Training, Validation, and Testing}
Each model, including the generic LSTM, \emph{mLSTM}, and its derivatives, was trained for $160000$ epochs on the same partition ($80\%$) of the dataset made from the combination of NREL and RAP data. Due to a large number of data points ($84096$), a training session would cycle through the training set less than 15 times, minimizing model overfitting. The validation set was also the same across all models, and it served as a quantifier of training progression. To ensure a fair comparison, the same test set was used across all models, which consisted of $1000$ data points from a non-trained partition of the dataset.
          
In order to measure model accuracy, the Normalized Mean Absolute Error (NMAE) was used. The NMAE and its variant, the Mean Absolute Error (MAE), is defined as follows, where $MAX(x)$ represents the wind farm capacity of 16 MW:
\begin{equation}
e_{MAE}(x', x) = \frac{1}{N} *\sum_{i=1}^{N}|x'(i)-x(i)|
\end{equation}
\begin{equation}
e_{NMAE}(x', x) = \frac{100\%}{MAX(x)}*\frac{1}{N} * \sum_{i=1}^{N}|x'(i)-x(i)|
\end{equation}
The NMAE allows test results to be better extrapolated to farms with different capacities, while the MAE gives a better quantification of true power error. Due to the non-convex nature of LSTM optimization, the initial weight states proved to be non-trivial in the final performance of the trained models. Thus, each model was trained 5 times and a margin of error (MOE) at $95\%$ was calculated. 

\subsection{Naive Ratio}
Throughout experimentation, it was observed that some model modifications had accuracies with non-statistically significant differences, yet step plots of their test sets revealed a wide range of naive character. Thus, to better compare models, a metric of naive character was created, which quantifies the step plot observations.  Under a pure persistence algorithm, the model predictions would be one timestep behind the truth, meaning that shifting the prediction backward one timestep would decrease the loss to zero. In the opposite case, where the model is perfectly accurate, shifting the prediction backward one timestep would increase the loss to a non-zero value. The models were neither of the boundary cases, but comparing a loss where the predictions are not time-shifted to a loss where the predictions are shifted one time-step into the past can give insight into the magnitude of naive behavior exhibited by the trained model. 

The back-shifted loss function is defined as follows, where subscript $t+1$ represents the back shift (the NMAE subscript was removed for conciseness):
\begin{equation}
e_{t+1}(x', x) = \frac{1}{N} * \sum_{i=1}^{N-1}|x'(i+1)-x(i)|
\end{equation}
If $e_{t+1}(x', x) < e_{t}(x', x)$ then the model is defined to have \emph{Naive Character}, and the presence of such character is placed on a continuous spectrum through the proposed \emph{Naive Ratio} (NR), which is defined as the following:
\begin{equation}
NR =\frac{ e_{t}(x', x)}{e_{t+1}(x', x)}, NR \in (0, \infty)
\end{equation}
A Naive Ratio score of anything greater than $1$ means that the model has some naive character, and a score of  $\infty$ indicates a perfectly naive model. A score of anything less than $1$ indicates that naive character takes a minority contribution to the model's behavior, and a perfectly accurate  model would have a Naive Ratio score of $0$.
\subsection{Control Models}
There were two control models used in experimentation that set benchmark values for the \emph{mLSTM} and its variants. As a baseline machine learning model, a generic LSTM with PCA was used, and as a baseline naive model, a persistence algorithm was used.
\section{Results and Analysis}
\subsection{Impact of Weather Data on Performance}
\begin{table}[htb]
\caption{Effect of Weather Data on Generic LSTM Model}
\begin{center}
\begin{tabular}{|m{1.6cm}|m{1.7cm}|m{2cm}|m{1.7cm}|}
 \hline
   &Power Data Only & With Weather & With Weather and PCA\\
 \hline
 NMAE (\%) & 2.409 & 1.649 & 1.130\\ 
 \hline
 95\% MOE & 0.450 & 0.231 & 0.212\\
 \hline
\end{tabular}
\label{Generic}
\end{center}
\end{table}

The forecast data worked as intended, with the forecast-supplemented model having a statistically significant improvement over the direct power model, and a PCA compression also having a statistically significant improvement over an uncompressed model (Table \ref{Generic}). 
\begin{table}[htb]
\caption{Effect of Data Type on Naive Ratio Score (Generic)}
\begin{center}
\begin{tabular}{|m{2.5cm}|m{1.5cm}|m{1.5cm}|m{1.5cm}|}
 \hline
    & Weather & Weather With PCA & Power Data Only \\
 \hline
 Naive Ratio Score & 1.050 & 1.353 & 1.145\\ 
 \hline
\end{tabular}
\label{dtnaive}
\end{center}
\end{table}

In Table \ref{dtnaive}, it is observed that the addition of uncompressed weather forecast data to the Generic LSTM models resulted in the decrease of naive character. Combined with the previously observed increase in accuracy, it can be concluded that weather forecast data is providing adequate context for wind power. 
\subsection{Model Modification Performance}
\begin{table}[htb]
\caption{Evaluation of Model Performance}
\begin{center}
\begin{tabular}{|m{0.7cm}|m{0.6cm}|m{0.85cm}|m{0.90cm}|m{0.85cm}|m{0.9cm}|m{1cm}|}
 \hline
   & Naive Ctrl. & LSTM w/PCA Ctrl.& \textbf{mLSTM} Direct Feed & \textbf{mLSTM} PCA & \textbf{mLSTM} Compr. & \textbf{mLSTM} CFIG + Peep \\
 \hline
 NMAE (\%) & 1.451 & 1.130 & 0.522 & 0.472 & 0.438 & 0.590\\
 \hline
95\% MOE & 0.262 & 0.212 & 0.038 & 0.038 & 0.031 & 0.038 \\
 \hline
 Naive Ratio & $\infty$ & 1.353 & 0.774 & 0.876 & 0.677 & 0.939 \\
 \hline
\end{tabular}
\label{modelresults}
\end{center}
\end{table}
Shown in Table \ref{modelresults} is a compilation of test results on the new models and controls. It is seen that the \emph{mLSTM}-based models had higher accuracies and lower Naive Ratio scores. This increase in performance is attributed to the bounding of cell state values and the expansion of cell state sizes, which keep the weights at a reasonable level and allow for greater storage of past information, respectively. 

\begin{figure}[htb] 
        \centering \includegraphics[width=\columnwidth]{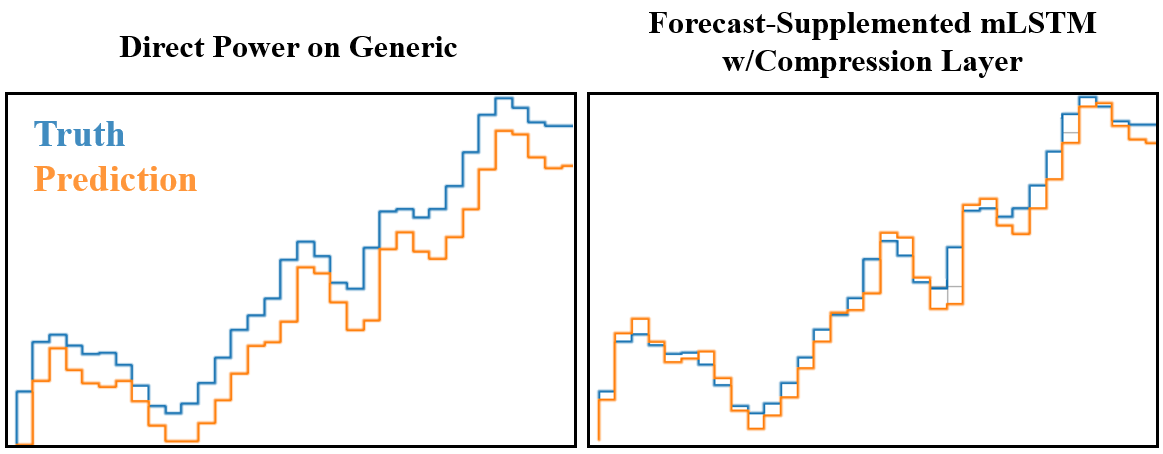}
                \caption{By adding forecast data and making some LSTM structural modifications, the naive character and prediction loss greatly decrease.}
                        \label{fig:goodplot}
                        \end{figure}

Among the variants, the compression layer addition showed the highest accuracy and lowest naive character. This improvement is clearly shown in Figure \ref{fig:goodplot}, which compares this version to the power-data-only generic LSTM on the same section of the test set. This is consistent with the observation made in Table \ref{Generic} that compression increases accuracy, which is caused by the lowered computational complexity and the isolation of trend-relevant data from noise. However, in observing the differences between the compression layer and PCA on the \emph{mLSTM}, the Naive Ratio score must be considered. The \emph{mLSTM} with PCA showed an increase in naive character from the unmodified \emph{mLSTM}, while the compression layer showed the opposite. Because an increase in naive character corresponds to a decrease in trend learning, it can be reasonably concluded that the PCA, while being able to capture data variation, was not able to optimally preserve trend-relevant data. However, the compression layer, trained with the model, was able to adapt its compression and learned to keep such trend-relevant data while still reducing complexity. The differences in naive character between the PCA and compression \emph{mLSTM} models shows that optimizing for variation may attenuate more relevant yet subtler data. The increased naive character with PCA is also observed on the Generic LSTM (Table \ref{dtnaive}). Taken holistically, it can be concluded that an adaptive compression method was a better feature encoding method for this application. 

It is worth noting that the NMAE of the persistence algorithm ($1.451\%$, Table \ref{modelresults}) was lower than that of a generic LSTM on uncompressed forecast data ($1.649\%$, Table \ref{Generic}). This shows that pure naive behavior has a deceptively low NMAE, which justifies its presence in some trained models. Thus, the naive ratio score is very important in disseminating false naive accuracy from real model accuracy. 

The results also show that combining CFIG and peephole on the \emph{mLSTM} decreased performance, both in terms of NMAE and Naive Ratio. However, the addition of the CFIG  allowed for a more uniform forget gate distribution after training, which indicates a stable convergence. The worse performance can be attributed to the Peephole addition. In an unlisted trial, a Peephole-only \emph{mLSTM} had an NMAE of $2.328\%$ and a Naive Ratio of $1.040$, both of which are significantly worse than an unmodified \emph{mLSTM}. Peepholes connect the cell states with the gates, which can cause unwanted gradient complications like local minima, which inhibits optimization. The addition of the CFIG modification decreased gate complexity and counteracted some negative effects of the Peephole, but the impact of the Peephole was still present in the final model.

\section{Conclusion, Limitations, and Future Development}
It was shown that using weather forecast data to supplement the power data is effective in increasing prediction accuracy, with an adaptive compression algorithm, cell regulation, and ReLU units serving to further increase accuracy and decrease naive character. The proposed \emph{mLSTM} models are able to outperform traditional LSTM models at a NMAE of $0.438\%$ (Table \ref{modelresults}) with the compression modification, or approximately 75 kW on a 16 MW farm. This improvement is also qualitatively observed in Figure \ref{fig:goodplot}. Such improvements, paired with future model developments, can be one step towards a more renewable energy-integrated future. 

However, it is worth noting that such findings are not guaranteed to generalize to all wind farms, whose behaviors vary with location and climate. As such, future developments include trials at multiple locations to measure the location consistency of the new models. The usage of region-based predictions will also be explored, which contain less noise than single-farm predictions. 

\section{Acknowledgements}
Special thanks to Zhenhua Liu and Joshua Comden of Stony Brook University for their mentorship and proofreading.

\bibliographystyle{unsrt}
\bibliography{ref}

\end{document}